# ADDRESSING THE KNAPSACK CHALLENGE THROUGH CULTURAL ALGORITHM OPTIMIZATION

Mohammad Saleh Vahdatpour

Department of Computer Science, College of Arts & Sciences,
Georgia State University, Atlanta, USA

*ABSTRACT*

*The "0-1 knapsack problem" stands as a classical combinatorial optimization conundrum, necessitating the selection of a subset of items from a given set. Each item possesses inherent values and weights, and the primary objective is to formulate a selection strategy that maximizes the total value while adhering to a predefined capacity constraint. In this research paper, we introduce a novel variant of Cultural Algorithms tailored specifically for solving 0-1 knapsack problems, a well-known combinatorial optimization challenge. Our proposed algorithm incorporates a belief space to refine the population and introduces two vital functions for dynamically adjusting the crossover and mutation rates during the evolutionary process. Through extensive experimentation, we provide compelling evidence of the algorithm's remarkable efficiency in consistently locating the global optimum, even in knapsack problems characterized by high dimensions and intricate constraints.*

## 1. INTRODUCTION

The 0-1 knapsack problem is one of the most studied combinatorial optimization problems. It contains a knapsack with a limited capacity and a set of N items, each with a weight (w) and a value (v) [30]. Solving this problem means finding a collection of items that the total weight of them is less than or equal to the knapsack capacity and the total value is as large as possible [1,11].

Suppose, *W* is the capacity of the knapsack (*W*>0), vector w = ($w_1$, $w_2$,..., $w_N$) and v=($v_1$,$v_2$,...,$v_N$) stand for the weight and the value of items, where $w_i$ > 0, $v_i$ > 0 , 1≤ i ≤ N.
Our aim is to find a set of $x_i$ where $x_i$ (*0 < i < N*) is equal to 0 or 1 which satisfies the constraints of Eq. (1) and Eq. (2) [2]:

$$\sum_{i=1}^{N} x_i w_i \le W \qquad (1)$$

$$max \sum_{i=1}^{N} x_i v_i \qquad (2)$$

If the i-th item is put into the knapsack then xi =1; otherwise xi =0.

Knapsack problem has numerous applications in theory and real world such as capital budgeting problems [3], loading problems [5], resource allocation [4] and project selection problems [6], also it can be found as a sub problem of the other general models.

Many methods have been developed to solve the knapsack problem: such as dynamic programming [7, 12], branch-and-bound approach [8], ant colony optimization [9, 13], particle swarm optimization [10, 14], simulated annealing [15], harmony search algorithm [16], amoeboid organism algorithm [17], schema-guiding evolutionary algorithm [20], soccer league algorithm [18] and so on.

In the real world, 0-1 knapsack problems usually have high dimension. Available algorithms lose their efficiency in solving such high dimensional 0-1 knapsack problems [2]. Hence, these algorithms are not suitable enough for real world scenarios. More research on this topic in required exploring more efficient and optimized solutions [2].

A Genetic Algorithm (GA) is a metaheuristic method for solving both constrained and unconstrained optimization problems based on a natural selection process that mimics biological evolution [28]. The algorithm repeatedly modifies a population of individual solutions.

Cultural Algorithms (CAs) are a population-based optimization technique inspired by the way societies preserve and share knowledge. In a CA, individuals within a population interact and adapt their knowledge through social learning, guided by a belief space. This belief space represents the shared knowledge of the population and plays a crucial role in shaping the algorithm's behavior. A cultural algorithm is a branch of evolutionary computation where in addition to the population component, there is a knowledge component that is called belief space [27]. Cultural algorithm can be seen as an extension of a conventional genetic algorithm.

In this paper, we propose a new approach for solving 0-1 knapsack problem based on CA. Our focus is to select items for belief space that result the best solution in minimum time. For this purpose, we introduce our situational and normative components. In addition, we employ GA for this purpose. In genetic algorithm, we tune parameters and fitness function for having the best solution.

The rest of the paper is organized as of the following: Section 2 reviews genetic algorithms and cultural algorithms. In Section 3 we introduce the proposed method for solving knapsack problem with cultural algorithms. We demonstrate the efficiency and accuracy of our method by comparing it to the existing methods in Section 4. Finally, Section 5 concludes the paper.

## 2. RELATED WORKS & BACKGROUND

In this section, we briefly review genetic and cultural algorithms

### 2.1. Genetic algorithm

Genetic Algorithm is a heuristic algorithm that mimics the process of natural selection [10]. This heuristic (aka, Meta heuristic) is used to generate useful solutions in optimization and search problems. Genetic algorithms are a subset of Evolutionary Algorithms (EA), which solve optimization problems using techniques inspired by natural evolution (e.g inheritance, mutation, selection, and crossover) [8].

In a genetic algorithm, a population of candidate solutions in an optimization problem evolves toward better solutions. This evolution happens through alteration and mutations in candidate solution's set of properties (usually called chromosomes or genotype). Binary or real vectors can represent candidate solutions (genotypes). Based on the problem, other representations can be employed.

Usually the iterative evolution process starts from a population includes individuals those are generated randomly. We will call the population in each iteration "a generation". The finesses of individuals are evaluated in each generation [13]. Fitness is usually defined by assessing the optimization the objective function of a problem. Regarding the results of this evaluation, the fittest individuals are selected stochastically, and individuals are altered by recombination and mutation functions to form a new generation. The new generation of candidate solutions is then evaluated and altered in the next iteration of the algorithm. The algorithm terminates when a satisfactory fitness value is achieved (optimization problem solved) or if a maximum number of generations have been produced.

The fitness function definition depends to the problem measures the excellence of a candidate solution. For instance, in the knapsack problem the objective is to maximize the total value of objects that can be put in a knapsack with a fixed capacity [29]. A representation of a solution can be a bit array, where each bit represents a different object and the value of the bit (0 or 1) shows whether or not the object is chosen to be in the knapsack [30]. Many of such representations are invalid, as the total size of objects may exceed the capacity of the knapsack. We can define the fitness of the solution to be the sum of values of all objects in the knapsack, if the representation is valid or 0 otherwise.

## 2.2. Cultural Algorithm

Cultural algorithms are a branch of evolutionary computation where in addition to the population component, there is a knowledge component called belief space. Cultural algorithms can be considered as an extension of a conventional genetic algorithm.

The best individuals of the population update the belief space after each iteration. Here, similar to genetic algorithms, the best individuals are selected using a fitness function that assesses the performance of each individual in population [19]. The belief space of a cultural algorithm can be separated to different categories. These categories represent different domains of knowledge that the population has about the search space. Normative and situational are example sub categories of belief space. Normative knowledge is a collection of desirable value ranges for the individuals in the population component. e.g., acceptable behavior for the agents in the population [14]. Situational knowledge includes specific examples of important events - e.g. successful/unsuccessful solutions. Figure. 1. shows the block diagram of a Cultural Algorithm.

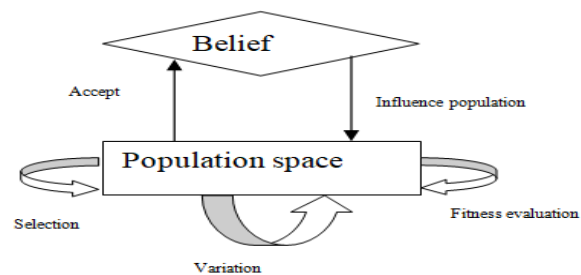

Figure 1. Block diagram of Cultural Algorithm

## 3. PROPOSED METHOD

In this section, we introduce our Cultural Algorithm and the procedure of building the belief space. The following sub sections describe each component separately. In formulating the Knapsack problem for our CA method, we make several key assumptions. Firstly, we assume a discrete representation of items, where each item is represented as a binary variable, indicating whether it is included (1) or not included (0) in the knapsack. Secondly, we assume a predefined capacity constraint for the knapsack. These assumptions are fundamental to the problem formulation in the context of our CA.

### 3.1. Fitness function

In Section 1, Eq. (1) and Eq. (2) formulate the fitness of a normal knapsack problem. In this paper, we propose a new approach for formulating the knapsack problem.

The cultural algorithm builds a chromosome of items for solving the knapsack problem. The procedure of chromosome building is described in the following:

Assume that the solution of the knapsack problem is represented by a binary sequence, where ith element is 1 if item i is put in the knapsack. The length of chromosome depends on the dimension of the problem. For example, for the problem called p5 (in Table 2), its chromosome represents the knapsack with 15 items [33].

Our new fitness function verifies the chromosome in construction process. When an item is unsuitable for the solution, its violation is calculated using Eq. (3) and a penalty is assigned to the old fitness. Our proposed fitness considers enough penalties for all the problems; in other words, we have an adaptive penalty for each situation. We indicate this with Eq. (4)

$$C = \max(\sum_{0}^{N}(x.*w) - W, \; 0) \quad (3)$$

$$Z = Z - \left(\frac{100}{d}\right) * C \quad (4)$$

Where variable Z in Eq. (3) and Eq. (4) is the value of fitness used in previous researches. The way of calculating Z is explained in Eq. (5). Variable d is used in Eq. (4) to address the high dimensionality problem. Value of variable d depends on the number of selected items, meaning higher dimension knapsack problem results in decreased penalty d.

$$d = 1 + \log\log N \quad (5)$$

Using this parameter capacitates fitness in high dimensional problems and ensures the approach does not lose efficiency.

### 3.2. Parameters Setting

Two important parameters of cultural algorithm are crossover rate (Pc) and mutation rate (Pm). We proposed a new formula to adjust the crossover rate. The focus of this formula is on iteration of the algorithm. At the beginning of the algorithm (means in initial iterations) exploration must be maximized and exploitation is in minimum value. In the middle iterations, rate of crossover and mutation can be equal for searching entirety of solution space. At the end, maximum exploitation is required for converging the algorithm to a solution. In this respect, in each iteration, the rate or probability for cross over and mutation operations is calculated by Eq. (6) and Eq. (7).

$$p_c = \min(\frac{p_c}{d} + \left[\frac{iter}{1000}\right] * 0.1, 1) \qquad (6)$$

$$p_m = 1 - p_c \qquad (7)$$

This formulation has been obtained by trial and error approach based on executing different experimentations.

### 3.3. Crossover and Mutation

Crossover and mutation are two basic operators of GA. Our crossover is a single-point crossover. A single crossover point is selected on both parents' chromosome strings. All data beyond that point in either chromosome string is swapped between the parent chromosomes [28]. The resulted chromosomes are the children. The parent selection mechanism is described in the next subsection.

The employed mutation in our algorithm is bit string mutation. Bit string mutation is the mutation of bit strings ensue through bit flips at random positions.

### 3.4. Belief Space

Belief space is an advantage of cultural algorithm compared to the genetic algorithm. In the following, we describe how we consider two functions, Accept, and Influence population [32], in Figure. 1.

1) Accept: At the first iteration, 10 percent of the best individuals in the population are selected for belief space creation. At other iterations, when an individual has a better fitness than the individuals in the belief space and up the 50 percent of thing selected for knapsack are different from individuals in belief space, belief space is updated with a new individual. Efficacy of belief space is in cross over.
2) Influence population: One of the parents for crossover is selected from belief space and the other parent is randomly selected from population. This change is like society. In society when an important people change the way of life other people of society accept this change without any question and efficacy of important people is very high.

### 4. EXPERIMENTAL RESULTS

In this section, we expand our comparison of the CA method with a broader range of optimization algorithms, including Genetic Algorithms, Particle Swarm Optimization, and Simulated Annealing. We present a comprehensive analysis of their performance on various benchmark problems, highlighting the strengths and weaknesses of each approach.

Previously, we focused on comparing our CA method with a specific set of algorithms. However, to provide a more holistic view of its performance, we have now included these additional algorithms in our comparative analysis.

Here, the performance of the GA and CA are studied by different experiments. All the computational experiments are conducted with Matlab R2013a in Intel® Core™i3 CPU M 350 @ 2.27GHz with 4 GB RAM system.

In order to comprehensively evaluate the performance of the proposed algorithm, we employ a combination of benchmark datasets and custom-generated problem instances. This approach allows us to assess the algorithm's effectiveness across a diverse range of problem scenarios, encompassing both synthetic instances designed to vary in complexity and dimensions, as well as real-world problem instances when applicable and available.

In all experiments, max iteration is set to 50 and the size of population is 100. The initial probability of crossover and mutation are set to Pc=0. 9 and Pm=0.1. the parameter setting has been shown in Table 1.

Table 1.  Parameter settings

| Description of parameter | Parameter name | Value |
| --- | --- | --- |
| Cross over rate | $P_c$ | 0.1 to 0.9 |
| Mutation rate | $P_m$ | $1-P_c$ |
| Iteration | iter | 50 to 2000 |
| Population size | Pop | 100 |
| Dimension | D | 5 to 1500 |

At first, we consider ten small 0-1 Knapsack problems which their details (contain weight of item w, value of item v, capacity of knapsack W and optimum solution) have been presented in Table 2 [2]. It should be noted that some studies have used these test cases for evaluating their methods. In [9], problems P1 and P2 are solved by an improved ant colony algorithm. P3 is used with a sequential combination tree algorithm in [19]. In [21] a greedy-policy-based algorithm examines P4. P5 is analyzed with the information about the search space landscape to search optimum solution in [22] and [23] has solved P6 in an approach similar to the shrinking boundary method P7 and P8 are considered with a nonlinear dimensionality reduction method in [25]. P9 is used by a DNA algorithm and P10 is used in [24]. It should be noted that problems Pi where i =3, 4... 10 have reached to optimum solution in the mentioned papers. The research of [2] presents a survey of the above studies and the results.

Table 3 reports the performance evaluation of the four algorithms Binary Gravitational Search Algorithm (BGSA) [2], CDGSA [2], GA, and CA (the proposed approach) on the 10 small problems described in Table 2. The table includes optimal solution and the best, worst, average, and median solution, the standard deviation (std.dev) and average total time (Avg. time) during 20 independent runs are presented in Table 3.

It should be noted that the solution of 35 means the total value of the selected items in the knapsack is equal to the number and a best solution means the optimum selection of items where their accumulative numbers yield the total number. Moreover, the worst solution of 83 means the minimum outcome of this knapsack is 83.

Table 3 shows that our approach has the faster convergence for the large problems in comparison with BGSA and CDGSA. BGSA finds local optimum solutions for many problems (such as Pi, i =2, 5, 8, 10).

Table 2. dimension, parameters and optimum solution of ten 0-1 knapsack problems (The parameter *w* and *v* are the weight and value of items, respectively; *W* is the capacity of the knapsack)

|  | Dim(N) | Parameter (w, W, v) | Optimum solution |
|---|---|---|---|
| $P_1$ | 10 | w=(95, 4, 60, 32, 23, 72, 80, 62, 65, 46)<br>W=269<br>v=(55, 10, 47, 5, 4, 50, 8, 61, 85, 87) | 295 |
| $P_2$ | 20 | w=(92, 4, 43, 83, 84, 68, 92, 82, 6, 44,32, 18, 56, 83, 25, 96, 70, 48, 14, 58)<br>W=878<br>v=(44, 46, 90, 72, 91, 40, 75, 35, 8, 54, 78, 40, 77, 15, 61, 17, 75, 29, 75, 63) | 1024 |
| $P_3$ | 4 | w=(6, 5, 9, 7)<br>W=20<br>v=(9, 11, 13, 15) | 35 |
| $P_4$ | 4 | w=(2, 4, 6, 7)<br>W=11<br>v=(6, 10, 12, 13) | 23 |
| $P_5$ | 15 | w=(56.358531, 80.874050, 47.987304, 89.596240, 74.660482, 85.894345, 51.353496, 1.498459, 36.445204, 16.589862, 44.569231, 0.466933, 37.788018, 57.118442, 60.716575)<br>W=375<br>v=(0.125126, 19.330424, 58.500931, 35.029145, 82.284005, 17.410810, 71.050142, 30.399487, 9.140294, 14.731285, 98.852504, 11.908322, 0.891140, 53.166295, 60.176397) | 481.0694 |
| $P_6$ | 10 | w=(30, 25, 20, 18, 17, 11, 5, 2, 1, 1)<br>W=60<br>v=(20, 18, 17, 15, 15, 10, 5, 3, 1, 1) | 52 |
| $P_7$ | 7 | w=(31, 10, 20, 19, 4, 3, 6)<br>W=50<br>v=(70, 20, 39, 37, 7, 5, 10) | 107 |
| $P_8$ | 23 | w=(983, 982, 981, 980, 979, 978, 488, 976, 972, 486, 486, 972, 972, 485, 485, 969, 966, 483, 964, 963, 961, 958, 959)<br>W=10000<br>v=(981, 980, 979, 978, 977, 976, 487, 974, 970, 485, 485, 970, 970, 484, 484, 976, 974, 482, 962, 961, 959,958, 857) | 9767 |
| $P_9$ | 5 | w=(15, 20, 17, 8, 31)<br>W=80<br>v=(33, 24, 36, 37, 12) | 130 |
| $P_{10}$ | 20 | w=(84, 83, 43, 4, 44, 6, 82, 92, 25, 83, 56, 18, 58, 14, 48, 70, 96, 32, 68, 92)<br>W=879<br>v=(91, 72, 90, 46, 55, 8, 35, 75, 61, 15, 77, 40, 63, 75, 29, 75, 17, 78, 40, 44) | 1025 |

In addition, we execute our approach on the other eight knapsack problems to evaluate its performance on the high dimensional problems. Weight and value of items are randomly generated in this problem and the number of items is as 100, 200, 300, 500, 800, 1000, 1200 and 1500 respectively, and the other parameter (corresponding knapsack's capacity) is 1100, 1500, 1700, 2000, 5000, 10000, 14000 and 16000 respectively. – $v_i$ (i =1,..., N)∼ rand(50, 100) – $w_i$ (i =1,..., N)∼ rand(5, 20)

288	Computer Science & Information Technology (CS & IT)All the reported results of this experiment have been obtained among 10 independent runs in which the best, worst, average, median solution and the standard deviation (std.dev) are presented in Table4.

Table 3. The results of the BDGSA, CDGSA, GA and CA on knapsack problems,
OP means the optimum solution

| P | OP | Method | Best | Worst | Average | Median | Avg. time (s) |
|---|---|---|---|---|---|---|---|
| $P_1$ | 295 | BGSA | 295 | 4 | 269.45 | 204.75 | 0.12 |
|  |  | CDGSA | 295 | 0 | 169.81 | 197.25 | 0.19 |
|  |  | GA | 295 | 294 | 294.73 | 295 | 0.31 |
|  |  | CA | 295 | 295 | 295 | 295 | 0.40 |
| $P_2$ | 1024 | BGSA | 972 | 83 | 892.05 | 631.75 | 3.14 |
|  |  | CDGSA | 1024 | 551 | 721.86 | 768 | 2.14 |
|  |  | GA | 1024 | 846 | 1012.34 | 1008 | 0.26 |
|  |  | CA | 1024 | 830 | 1004.63 | 993.52 | 0.35 |
| $P_3$ | 35 | BGSA | 35 | 0 | 34.35 | 29 | 0.03 |
|  |  | CDGSA | 35 | 0 | 24.52 | 28 | 0.05 |
|  |  | GA | 35 | 35 | 35 | 35 | 0.29 |
|  |  | CA | 35 | 35 | 35 | 35 | 0.31 |
| $P_4$ | 23 | BGSA | 23 | 0 | 21.55 | 17.75 | 0.04 |
|  |  | CDGSA | 23 | 0 | 14.96 | 18 | 0.04 |
|  |  | GA | 23 | 23 | 23 | 23 | 0.27 |
|  |  | CA | 23 | 23 | 23 | 23 | 0.29 |
| $P_5$ | 481.0694 | BGSA | 475.4784 | 59.79 | 410.81 | 321.77 | 1.84 |
|  |  | CDGSA | 481.0694 | 26.76 | 249.88 | 282.23 | 1.14 |
|  |  | GA | 481.0694 | 398 | 469.78 | 451.50 | 0.25 |
|  |  | CA | 481.0694 | 402.05 | 473.41 | 465.52 | 0.26 |
| $P_6$ | 51 | BGSA | 52 | 1 | 47.06 | 41.5 | 0.21 |
|  |  | CDGSA | 52 | 0 | 37.12 | 38 | 0.26 |
|  |  | GA | 52 | 51 | 51.85 | 52 | 0.24 |
|  |  | CA | 52 | 52 | 52 | 52 | 0.26 |
| $P_7$ | 107 | BGSA | 107 | 0 | 94.35 | 74 | 0.03 |
|  |  | CDGSA | 107 | 5 | 66.03 | 80.5 | 0.04 |
|  |  | GA | 107 | 107 | 107 | 107 | 0.30 |
|  |  | CA | 107 | 107 | 107 | 107 | 0.30 |
| $P_8$ | 9767 | BGSA | 9758 | 2316 | 9740.8 | 8445 | 4.21 |
|  |  | CDGSA | 9767 | 2316 | 8971.56 | 9254.1 | 3.26 |
|  |  | GA | 9767 | 9725 | 9749.13 | 9747.6 | 0.32 |
|  |  | CA | 9767 | 9732 | 9753.64 | 9749.2 | 0.36 |
| $P_9$ | 130 | BGSA | 130 | 0 | 116.35 | 98.25 | 0.03 |
|  |  | CDGSA | 130 | 0 | 94.10 | 106 | 0.04 |
|  |  | GA | 130 | 130 | 130 | 130 | 0.25 |
|  |  | CA | 130 | 130 | 130 | 130 | 0.26 |
| $P_{10}$ | 1025 | BGSA | 953 | 116 | 875.25 | 633.25 | 3.29 |
|  |  | CDGSA | 1025 | 116 | 749.51 | 832.92 | 3.38 |
|  |  | GA | 1025 | 934 | 1004.97 | 1016 | 0.30 |
|  |  | CA | 1025 | 1002 | 1015.67 | 1014.58 | 0.36 |

From Table 4 it is clear that the proposed method is faster to solve the high dimensional 0-1 knapsack problems in comparison to BGSA and CDGSA. Another observation that can be deduced from Table 4 is that the proposed method has gained better results in solving some of the high dimensional 0-1 knapsack problems.

Table 4.   The results of GA, CA (proposed method), DGSA and CDGSA on eight random large knapsack problems

| P | Method | Best | Worst | Average | Median |
|---|--------|------|-------|---------|--------|
| $P_{11}$ | DGSA | 7029 | 6034 | 6602.01 | 6715 |
| | CDGSA | 7245 | 5172 | 6449.20 | 6581 |
| | GA | 7142 | 5401 | 6521.73 | 6645 |
| | CA | 7230 | 5358 | 6543.74 | 6613 |
| $P_{12}$ | DGSA | 11024 | 9821 | 10429.58 | 10598 |
| | CDGSA | 11168 | 8267 | 11006.5 | 11063 |
| | GA | 11752 | 9498 | 11375.23 | 11401 |
| | CA | 11985 | 9462 | 11451.51 | 11498 |
| $P_{13}$ | DGSA | 13892 | 12957 | 13379.2 | 13420 |
| | CDGSA | 14025 | 12730 | 13602.09 | 13732 |
| | GA | 14017 | 11245 | 13704.61 | 13808 |
| | CA | 14043 | 11176 | 13729.54 | 13784 |
| $P_{14}$ | DGSA | 21891 | 20063 | 20957.14 | 21164 |
| | CDGSA | 27451 | 22173 | 25903.55 | 26508 |
| | GA | 27516 | 21542 | 25781.46 | 26461 |
| | CA | 27421 | 22067 | 25642.84 | 26553 |
| $P_{15}$ | DGSA | 47213 | 45810 | 46431 | 46605 |
| | CDGSA | 55048 | 47204 | 52761.37 | 53892 |
| | GA | 54634 | 45006 | 51637.76 | 52953 |
| | CA | 56842 | 44529 | 52387.63 | 53532 |
| $P_{16}$ | DGSA | 70825 | 68941 | 69730.12 | 69940 |
| | CDGSA | 73698 | 69752 | 70454.38 | 71653 |
| | GA | 73438 | 65645 | 69893.45 | 71748 |
| | CA | 72983 | 67632 | 70532.68 | 70985 |
| $P_{17}$ | DGSA | 89617 | 86752 | 88013.75 | 88351 |
| | CDGSA | 90139 | 86984 | 89452.64 | 89953 |
| | GA | 91834 | 54738 | 86375.91 | 87840 |
| | CA | 92745 | 60536 | 87753.52 | 88421 |
| $P_{18}$ | DGSA | 109541 | 107203 | 108429.8 | 108870 |
| | CDGSA | 111964 | 108927 | 110726.79 | 111248 |
| | GA | 112753 | 102231 | 109893.63 | 111051 |
| | CA | 115948 | 107255 | 111735.86 | 112372 |

In another experiment, we have selected four medium and large problems for evaluating the performance of our Cultural Algorithm. We have reported the convergence diagram of this experiment in Figure. 2.

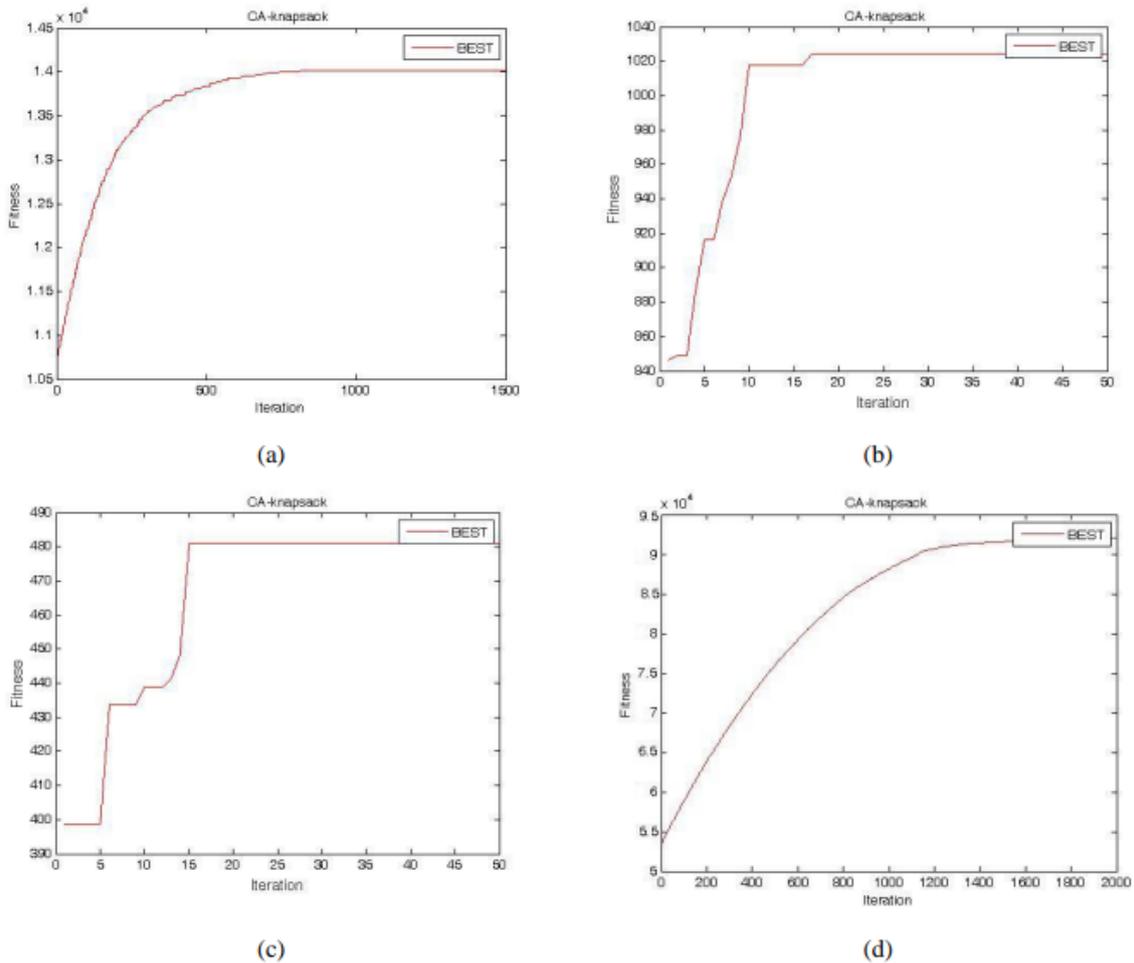

Figure. 2. Convergence diagram for different size knapsack problems(a) For large problem p13. (b) For medium problem p2 form table 3. (c) For medium problem p5 from table 3. (d) For large problem p17.

## 5. CONCLUSION

Knapsack problem is an NP-hard problem. Several algorithms have been developed to solve this problem in this paper, we improve GA by adjusting the rate of mutation and crossover in the evolution of the population. In addition, we employ the proposed cultural algorithm by constructing a belief space and considering its influence on the population. We apply both GA and CA to solve 0-1 knapsack problems. Experimental results show the superior efficiency and accuracy of the proposed method compared to the other algorithms.

While the CA method shows promise in solving the Knapsack problem, it is essential to acknowledge its limitations. One limitation is that the CA's performance may be sensitive to its parameter settings, requiring careful tuning. Additionally, the CA may struggle with high-dimensional problem spaces or problems with complex constraints. Future research should explore strategies to mitigate these limitations and improve the algorithm's robustness.

## AUTHORS

**Mohammad Saleh Vahdatpour** is a Ph.D. student in Computer Science at Georgia State University, USA.  He holds a Master's degree in Computer Science from the University of Tehran, Iran. His research interests include Statistic Machine Learning and Machine Vision, and he is committed to advancing the field through innovative research.

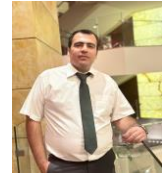